\pdfoutput=1
\documentclass[11pt]{article}
\usepackage[final]{acl}

\usepackage{soul}
\usepackage{times}
\usepackage{latexsym}
\usepackage{graphicx}
\usepackage{multirow}
\usepackage{multicol}
\usepackage{amsmath}
\usepackage{amsfonts}
\usepackage{amssymb}
\usepackage{algorithm}
\usepackage{algpseudocode}
\usepackage{booktabs}
\usepackage{booktabs}
\usepackage{multirow}
\usepackage{graphicx}
\usepackage{subcaption}
\usepackage{tcolorbox}
\usepackage{xcolor}
\usepackage{hyperref}
\usepackage{anyfontsize}
\usepackage{colortbl}
\usepackage{arydshln} 

\colorlet{lightyellow}{yellow!50}
\colorlet{lightcyan}{cyan!50}

\newcommand{\highlightyellow}[1]{\sethlcolor{lightyellow}\hl{#1}}
\newcommand{\highlightyelloww}[1]{\sethlcolor{yellow}\hl{#1}}
\newcommand{\highlightcyan}[1]{\sethlcolor{lightcyan}\hl{#1}}
\newcommand{\highlightpink}[1]{\sethlcolor{pink}\hl{#1}}

\usepackage[T1]{fontenc}
\usepackage[utf8]{inputenc}
\usepackage{microtype}
\usepackage{inconsolata}
\usepackage{graphicx}

\title{GuRE:Generative Query REwriter for Legal Passage Retrieval}

\author{Daehui Kim$^{1,2}$, Deokhyung Kang$^1$, Jonghwi Kim$^1$, Sangwon Ryu$^1$, Gary Geunbae Lee$^1{^3}$ \\
  $^1$Graduate School of Artificial Intelligence, POSTECH, Republic of Korea \\
  $^2$AI Future Lab, KT, Republic of Korea \\
  $^3$Department of Computer Science and Engineering, POSTECH, Republic of Korea \\
  \texttt{\{andrea0119, deokhk, jonghwi.kim, ryusangwon, gblee\}@postech.ac.kr} \\}

\begin{document}
\maketitle
\begin{abstract}
Legal Passage Retrieval (LPR) systems are crucial as they help practitioners save time when drafting legal arguments.
However, it remains an underexplored avenue.
One primary reason is the significant vocabulary mismatch between the query and the target passage.
To address this, we propose a simple yet effective method, the \textbf{G}enerative q\textbf{u}ery \textbf{RE}writer \textbf{(GuRE)}.
We leverage the generative capabilities of Large Language Models (LLMs) by training the LLM for query rewriting.
\textit{"Rewritten queries"} help retrievers to retrieve target passages by mitigating vocabulary mismatch.
Experimental results show that GuRE significantly improves performance in a retriever-agnostic manner, outperforming all baseline methods.
Further analysis reveals that different training objectives lead to distinct retrieval behaviors, making GuRE more suitable than direct retriever fine-tuning for real-world applications. 
Codes are avaiable at \href{https://github.com/daehuikim/GuRE}{github.com/daehuikim/GuRE}.
\end{abstract}

\newtcolorbox{verbatimbox}{
  colback=black!4, 
  colframe=black!4, 
  arc=1pt, 
  boxrule=0.5pt, 
  fontupper=\ttfamily 
}

\section{Introduction}
Recent advancements in information retrieval have enhanced legal tasks \cite{zhu2024largelanguagemodelsinformation,LAI2024181, tu2023artificial}. 
Most studies have focused on retrieving legal cases \cite{10.1145/3404835.3463250,10.1145/3626772.3657887,hou2024clercdatasetlegalcase,deng-etal-2024-element,deng-etal-2024-learning,gao-etal-2024-enhancing-legal} to address the challenge of retrieving relevant cases from the vast amount of documents. 
While automatic case retrieval systems are advancing, practitioners still spend significant time searching for relevant cases during argument drafting \cite{legal-challenge}.
One reason for this is that  cases frequently address multiple legal issues, so retrieved cases may be relevant overall but not necessarily contain passages that align with the specific argument being drafted.
As a result, practitioners often need to manually sift through lengthy documents to locate the specific passages for their argument.
Therefore, Legal Passage Retrieval (LPR) is crucial for extracting fine-grained information at the passage level, which helps reduce the time spent on legal research and lowers the costs associated with argument drafting.

\begin{figure}
    \centering
\includegraphics[width=0.95\columnwidth]{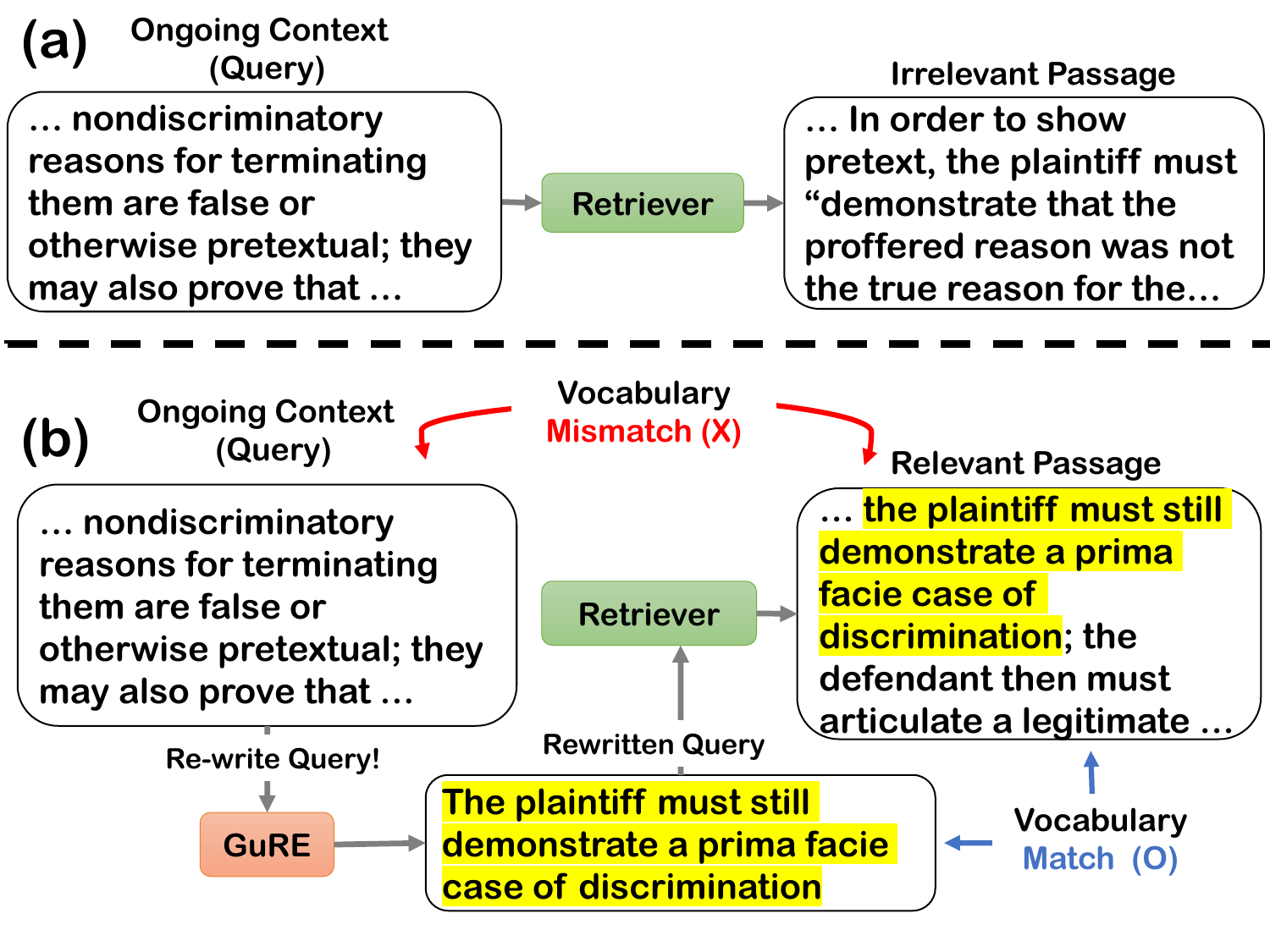} \vspace{-7pt}
    \caption{(a) Retriever fails to retrieve the target passage using an original query.
(b) GuRE rewrites the query before retrieval. Overlapping context between the \textit{"rewritten query"} and the target passage is in \highlightyelloww{yellow}.}
    \label{fig:1}
\end{figure}

Despite its importance, however, LPR remains underexplored, showing suboptimal performances even with fine-tuned retrievers
 \cite{mahari-etal-2024-lepard}.
One of the primary reasons for this is the significant vocabulary mismatch between the ongoing context (query) and the target passage \cite{nogueira2019document,feng-etal-2024-legal,mahari-etal-2024-lepard,hou2024clercdatasetlegalcase}.
In legal texts, queries frequently use terms that differ from those in the target passage, hindering retrievers from matching relevant passages \cite{valvoda-etal-2021-precedent}.
Figure \ref{fig:1} provides an example of the impact of vocabulary mismatch.

To address this challenge, we tried to modify the query to mitigate the vocabulary mismatch via the existing query expansion methods \cite{wang-etal-2023-query2doc,jagerman2023query}.
However, a substantial gap between the query and the target passage remained.
To bridge this gap, we propose a simple yet effective method, the \textbf{G}enerative q\textbf{u}ery \textbf{RE}writer \textbf{(GuRE)}.
We aim to enable Large Language models (LLMs) to leverage legal domain-specific knowledge better to rewrite queries with a mitigated vocabulary gap.
Specifically, We train LLMs to generate legal passages based on a query, which then serves as the \textit{"rewritten query"} for retrievers.
At retrieval time, we employ a \textit{"rewritten query"} with lower vocabulary mismatch as the query for the retriever, as shown in (b) of Figure \ref{fig:1}.

Experimental results demonstrate that retrieving using \textit{"rewritten queries"} from GuRE leads to a significant performance improvement in a retriever-agnostic manner, even surpassing direct retriever fine-tuning.
Our analysis reveals that adapting GuRE for LPR can be more suitable for real-world applications than direct retriever fine-tuning regarding their different training objectives.

Our contributions include a simple yet effective \textbf{domain-specific query rewriting method} to address the vocabulary mismatch problem in LPR.
We also analyze \textbf{why retriever fine-tuning leads to suboptimal performance in LPR, linking it to its training objective.}

\section{Method: GuRE}
We introduce GuRE, a simple yet effective method for mitigating the underlying vocabulary mismatch in LPR.
Unlike existing query expansion methods, which add additional information to the query, GuRE is designed to rewrite the query directly.
We train the LLM on a dataset of $InstructionPrompt_{q,p_q}$, where $q$ is \{Context\} and $p_q$ is \{Passage\} (Figure \ref{fig:2}).
Given a sequence of tokens \( (t_1, ..., t_N) \) from an $InstructionPrompt_{q,p_q}$, the LLM learns to predict each token \( t_i \) in auto-regressive manner by optimizing the Cross-Entropy loss: \vspace{-5pt}
\[
\mathcal{L} = - \sum_{i=1}^{N} \log P(t_i | t_{<i}; \theta)
\] 
Where \( P(t_i | t_{<i}; \theta) \) is the probability assigned by the model to the token \( t_i \) given previous tokens. 
$\theta$  is the parameters of the LLM.
Once trained, GuRE re-write the queries using the $InstructionPrompt_{q}$ excluding the \{Passage\} from Figure \ref{fig:2}.

\begin{figure}[t]
\centering
\small
\textbf{Instruction Prompt}
\begin{verbatimbox}
You are a helpful assistant specializing in generating legal passages that naturally align with the preceding context.

Based on the given preceding context, please generate a legal passage that is coherent, relevant, and contextually appropriate.

\#\#\# Preceding Context : \{Context\}

\#\#\# Legal Passage : \{Passage\}
\end{verbatimbox} \vspace{-7pt}
\caption{Instruction prompt for GuRE.}
\label{fig:2}
\end{figure}
\vspace{-7pt}

\begin{table*}[t]
\centering
\resizebox{0.9\textwidth}{!}{%
\begin{tabular}{@{}llccccccccc@{}}
\toprule
\multicolumn{2}{l}{} & \multicolumn{3}{c|}{10K}  & \multicolumn{3}{c|}{20K}  & \multicolumn{3}{c}{50K}  \\ \hline
\multicolumn{1}{c}{Type}    &Method& R @ 1     & R @ 10    & \multicolumn{1}{c|}{nDCG @ 10}& R @ 1    & R @ 10   & \multicolumn{1}{c|}{nDCG @ 10}& R @ 1    & R @ 10   & nDCG @ 10\\ \hline
\multirow{4}{*}{Sparse}     & \small BM25    & 9.91     & 28.19    & \multicolumn{1}{c|}{15.33}    & 8.81     & 24.51    & \multicolumn{1}{c|}{15.91}    & 7.37     & 20.83    & 13.41    \\
    &   \small BM25 + Q2D   & 10.23    & 34.99    & \multicolumn{1}{c|}{21.15}    & 8.55     & 28.89    & \multicolumn{1}{c|}{17.46}    & 6.57     & 22.58    & 13.63    \\
    &   \small BM25 + Q2D-CoT     & 11.13    & 35.96    & \multicolumn{1}{c|}{22.22}    & 9.29     & 30.37    & \multicolumn{1}{c|}{18.59}    & 7.48     & 24.03    & 14.81    \\
    &   \small BM25 + GuRE  & \textbf{$\text{34.88}^\dagger$} & \textbf{$\text{62.20}^\dagger$} & \multicolumn{1}{c|} {\textbf{$\text{47.69}^\dagger$}} & \textbf{$\text{28.39}^\dagger$} & \textbf{$\text{52.63}^\dagger$} & \multicolumn{1}{c|} {\textbf{$\text{39.69}^\dagger$}} & \textbf{$\text{19.41}^\dagger$} & \textbf{$\text{39.20}^\dagger$} & \textbf{$\text{28.46}^\dagger$} \\ \midrule

\multirow{10}{*}{Dense}& \small DPR     & 1.99     & 6.39     & \multicolumn{1}{c|}{3.92}     & 1.74     & 5.49     & \multicolumn{1}{c|}{3.39}     & 1.42     & 4.36     & 2.71     \\
    &   \small DPR + Q2D   & 1.92     & 7.22     & \multicolumn{1}{c|}{4.22}     & 1.54     & 6.07     & \multicolumn{1}{c|}{3.46}     & 1.08     & 4.08     & 2.39     \\
    &   \small DPR + Q2D-CoT     & 2.3& 7.98     & \multicolumn{1}{c|}{4.78}     & 1.92     & 6.84     & \multicolumn{1}{c|}{4.05}     & 1.35     & 4.86     & 2.86     \\
    &   \small DPR + GuRE  & \textbf{$\text{32.07}^\dagger$} & 49.74  & \multicolumn{1}{c|}{\textbf{$\text{40.68}^\dagger$}} & \textbf{$\text{26.35}^\dagger$} & 41.96 & \multicolumn{1}{c|}{\textbf{$\text{33.77}^\dagger$}} & \textbf{$\text{16.47}^\dagger$} & 30.63 & \textbf{$\text{23.20}^\dagger$} \\ \cmidrule(l){2-11} 
& \small DPR-FT  & 14.09   & \textbf{$\text{50.97}^\dagger$}    & \multicolumn{1}{c|}{30.31}    & 11.28    & \textbf{42.59}    & \multicolumn{1}{c|}{24.90}    & 8.23     & \textbf{31.07}    & 18.13    \\ \cmidrule(l){2-11}
    & \small ModernBert    & 7.11     & 22.47    & \multicolumn{1}{c|}{13.94}    & 6.04     & 19.16    & \multicolumn{1}{c|}{11.90}    & 4.94     & 15.24    & 9.58     \\
    &   \small ModernBert + Q2D   & 6.67     & 24.95    & \multicolumn{1}{c|}{14.67}    & 5.65     & 20.64    & \multicolumn{1}{c|}{12.19}    & 4.09     & 15.47    & 9.09     \\
    &   \small ModernBert + Q2D-CoT     & 7.47     & 26.46    & \multicolumn{1}{c|}{15.86}    & 6.47     & 21.99    & \multicolumn{1}{c|}{13.32}    & 4.90     & 16.96    & 10.22    \\
    &   \small ModernBert + GuRE  & \textbf{$\text{33.14}^\dagger$} & \textbf{$\text{60.24}^\dagger$} & \multicolumn{1}{c|}{\textbf{$\text{45.86}^\dagger$}} & \textbf{$\text{26.36}^\dagger$} & \textbf{$\text{51.34}^\dagger$} & \multicolumn{1}{c|}{\textbf{$\text{38.19}^\dagger$}} & \textbf{$\text{17.44}^\dagger$} & \textbf{$\text{37.89}^\dagger$} & \textbf{$\text{26.83}^\dagger$} \\ \cmidrule(l){2-11}
    & \small ModerBert-FT  & 14.12    & 51.34    & \multicolumn{1}{c|}{30.50}    & 11.51    & 42.31    & \multicolumn{1}{c|}{24.49}    & 8.75     & 31.81    & 18.80    \\ \bottomrule
\end{tabular}%
} \vspace{-5pt}
\caption{Evaluation results for various retrieval methods with different numbers of target passages ($N$k). The best performance for each retriever, across all metrics, is highlighted in \textbf{bold}. $\dagger$ denotes a statistically significant improvement (paired $t$-test, $p < 0.01$) over the best-performing method excluding those marked in \textbf{bold}.
}
\label{tab:1}
\end{table*}
\vspace{-3pt}

\section{Experiments} \vspace{-5pt}
\subsection{Task Description}
LPR involves retrieving the most relevant passage \( p_{q} \) based on an ongoing context \( q \), where $q$ serves as the query for the retriever. Given a set of candidate passages \( P_{collection} = \{p_1, \dots, p_n\} \), our goal is to identify \( p_q \in P_{collection} \) that can support \( q \) during the legal document drafting.

\subsection{Baselines}
Due to the absence of prior research on LPR, we compare GuRE with strong baselines as follows.
\paragraph{Query Expansion.} 
Query2Doc \textbf{(Q2D)} \cite{wang-etal-2023-query2doc} generates a pseudo-passage via few-shot prompting and concatenates it with the original query to form an expanded query.
Query2Doc-CoT \textbf{(Q2D-CoT)} \cite{jagerman2023query} extends Query2Doc by generating reasoning steps while producing the pseudo-passage.
We employ GPT-4o-mini \cite{openai2024gpt4technicalreport} for Q2D and Q2D-CoT.
Detailed settings are in the Appendix \ref{appendixb}.

\paragraph{Fine-Tuning} Since we train the LLM to build GuRE, we include retriever fine-tuning in the baseline to analyze the effectiveness of the training strategy.
We train the retrievers using Multiple Negatives Ranking Loss \cite{henderson2017efficient} by following \citeauthor{mahari-etal-2024-lepard}, maximizing the model similarity for a positive sample while minimizing similarity for other samples within a batch.
Details about baselines are in Appendix \ref{appendix}.

\subsection{Dataset}  
We use LePaRD \cite{mahari-etal-2024-lepard}, a representative large-scale legal passage retrieval dataset for U.S. federal court precedents.
It contains metadata along with ongoing context \( q \) and its corresponding cited target passage \( p_q \). 
The dataset includes three versions varying the size of the candidate passage pool, namely 10K, 20K, and 50K.
Each version consists of 1.9M, 2.5M, and 3.5M data points, respectively.
We use 90\% of each version for fine-tuning retrievers and training GuRE.
To ensure efficiency and reliability given the large scale of the dataset, we sample 10,000 data points three times from the remaining 10\% of the data and report the average over three trials.
Details of statistics are in the Appendix \ref{appendixa}.

\vspace{-5pt}
\subsection{Models} \vspace{-7pt}
We select SaulLM-7B \cite{colombo2024saullm} as the backbone model for GuRE, as it is pre-trained on a legal domain corpora. 
We also compare Llama3.1-8B \cite{grattafiori2024llama3herdmodels} and Qwen2.5-7B \cite{qwen2025qwen25technicalreport} as backbone models to assess the generalization of our approach across different backbone models.
The investigation of backbone model selection is provided in the Appendix \ref{appendixbackbone}.

We use BM25 \cite{robertson2009probabilistic}, DPR \cite{karpukhin-etal-2020-dense} and ModernBert \cite{warner2024smarterbetterfasterlonger} for retrievers.
More details about the retrievers are provided in Appendix \ref{appendixc}.

\vspace{-7pt}
\section{Results} \vspace{-7pt}
Table \ref{tab:1} reveals that adapting GuRE for query rewriting significantly improves retrieval performance across different methods and passage sizes. 
Notably, applying GuRE to BM25 results in a performance gain of 32.96 (15.33 $\rightarrow$ 47.69) in nDCG@10 for the 10K dataset. 
This significant improvement is consistent across all data versions (10K, 20K, 50K) and retrieval methods, highlighting the retriever-agnostic effectiveness of GuRE.

In contrast, other baseline methods yield suboptimal performance gains, falling short of the improvements by GuRE.
Q2D achieves the lowest performance gain, suggesting that the \textbf{few-shot prompting strategy struggles to address the underlying challenges in tasks requiring domain-specific knowledge.} 
Furthermore, retriever fine-tuning does not provide retrievers with the same level of performance as GuRE.
This indicates that \textbf{mitigating vocabulary mismatch is significantly more effective than training the retrievers.}

\vspace{-7pt}
\begin{table}[ht]
\centering
\resizebox{0.90\columnwidth}{!}{%
\begin{tabular}{@{\centering}p{1.5cm}ccccc@{}}
\toprule
        & BLEU & ROUGE-L & BertScore-F & Words \\ \midrule
\small Target & -    & -     & -         & 50.21           \\ \midrule[0.1pt]
\small Query   & 5.75    & 18.98     & 75.61         & 123.99           \\
\small Q2D     & 8.56    & 19.19     & 78.6         & 88.19           \\
\small Q2D-CoT  & 11.86    & 27.28     & 80.1         & 36.28           \\
\small GuRE    & \textbf{59.43}    & \textbf{67.62}     & \textbf{90.92}         & \textbf{50.90}           \\ \bottomrule
\end{tabular}%
} \vspace{-7pt}
\caption{Quantitative evaluation of pseudo-passages (Q2D, Q2D-CoT) and \textit{"rewritten query"} (GuRE) between target passages on the 10K test set.}
\label{tab:3}
\end{table} \vspace{-7pt}

\begin{table*}[t]
\centering
\resizebox{0.9\textwidth}{!}{%
\begin{tabular}{@{}p{0.08\textwidth}p{0.92\textwidth}@{}}
\toprule
\small Target Passage & \small Likelihood of Confusion.
The ultimate inquiry in most actions for false designation of origin, as with \highlightyellow{actions for trademark infringement, is whether there exists a “likelihood that an appreciable number of ordinarily prudent purchasers [will] be misled, or indeed simply confused, as to the source of the goods in question.}
 \\ \hline \midrule
\small Query   & \small   See Thompson Medical Co., Inc. v. Pfizer, Inc., 753 F.2d 208, 213 (2 Cir.1985) (quoting Mushroom Makers, Inc. v. R.G. Barry Corp., 580 F.2d 44, 47 (2 Cir.1978), cert. denied, 439 U.S. 1116 (1979)) (“   \\ \midrule
\small Q2D     & \small \highlightpink{[a] plaintiff's burden in establishing liability requires more than mere speculation; the evidence must be sufficient to show that the defendant's conduct was a substantial factor in bringing about the harm." This standard underscores the necessity for plaintiffs to provide concrete evidence linking the defendant's actions to the alleged damages, rather than relying on generalized assertions or conjectures.}  \\ \midrule
\small Q2D-CoT  & \small the standard for establishing \highlightyellow{trademark infringement is whether there is a likelihood of} confusion among consumers as to the source of the goods or services, focusing on factors such as strength of the mark, proximity of the goods, similarity of the marks, evidence of actual confusion,\highlightpink{ and the defendant’s intent in adopting its mark.}   \\ \midrule
\small GuRE   & \small 
II It is well settled that the crucial issue in an \highlightyellow{action for trademark infringement or unfair competition is whether there is any likelihood that an appreciable number of ordinarily prudent purchasers are likely to be misled, or indeed simply confused, as to the source of the goods in question.}
\\ \bottomrule
\end{tabular}%
}
\caption{Case study about generated pseudo-passage and \textit{"rewritten query"}. \highlightyellow{Yellow} indicates parts similar to the target passage, while \highlightpink{pink} marks \textit{"distractor"} that can mislead retrievers into wrong passages.}
\label{tab:4}
\end{table*}

\vspace{-7pt}
\section{Analyses} \vspace{-7pt}
\subsection{Rewritten Query Evaluation} \vspace{-7pt}
We analyze the generated context using various methods to investigate how effectively vocabulary mismatch is mitigated.
Table \ref{tab:3} shows a quantitative evaluation of pseudo-passages (Q2D, Q2D-CoT) and \textit{"rewritten queries"} (GuRE) against target passages on the 10K test set.
The highest metric values reflect the \textbf{high lexical similarity between GuRE's \textit{"rewritten queries"} and target passages,} while pseudo-passages from Q2D and Q2D-CoT struggle to mitigate the lexical gap.

Additionally, we find that the \textit{"rewritten query"} generated by GuRE contains semantically similar legal context to the target passage (Table \ref{tab:4}).
For example, GuRE successfully generates phrases like \highlightyellow{"action for trademark infringement"}. 
In contrast, pseudo-passages from Q2D are \highlightpink{mostly irrelevant}, and while Q2D-CoT generates some relevant context like \highlightyellow{"trademark infringement"}, it also produces irrelevant context such as \highlightpink{"defendant's intent in adopting its mark"}.
These results show that \textbf{domain-specific training outperforms few-shot prompting in mitigating vocabulary mismatch.}
More case-studies are in the appendix \ref{appendixg}.

\begin{table}[ht]
\centering
\resizebox{\columnwidth}{!}{%
\begin{tabular}{@{}
>{\columncolor[HTML]{FFFFFF}}c 
>{\columncolor[HTML]{FFFFFF}}c 
>{\columncolor[HTML]{FFFFFF}}c 
>{\columncolor[HTML]{FFFFFF}}c @{}}
\toprule
                          & \multicolumn{3}{c}{\cellcolor[HTML]{FFFFFF}10K Cases}                                                                 \\ \midrule
                          & R@1                                   & R@10                                  & nDCG@10                               \\ \midrule
ModernBert                & {\color[HTML]{333333} 7.11}           & {\color[HTML]{333333} 22.47}          & {\color[HTML]{333333} 13.94}          \\
GuRE (10K) + ModerBert    & {\color[HTML]{2C3A4A} 16.42}          & {\color[HTML]{2C3A4A} 39.02}          & {\color[HTML]{2C3A4A} 26.58}          \\
GuRE (100 K) + ModernBert & {\color[HTML]{2C3A4A} \textbf{20.62}} & {\color[HTML]{2C3A4A} \textbf{45.98}} & {\color[HTML]{2C3A4A} \textbf{32.31}} \\ \midrule
                          & \multicolumn{3}{c}{\cellcolor[HTML]{FFFFFF}{\color[HTML]{333333} 20K Cases}}                                          \\ \midrule
ModernBert                & {\color[HTML]{333333} 6.04}           & {\color[HTML]{333333} 19.16}          & {\color[HTML]{333333} 11.9}           \\
GuRE (10K) + ModerBert    & {\color[HTML]{2C3A4A} 12.06}          & {\color[HTML]{2C3A4A} 31.29}          & {\color[HTML]{2C3A4A} 20.81}          \\
GuRE (100 K) + ModernBert & {\color[HTML]{2C3A4A} \textbf{15.35}} & {\color[HTML]{2C3A4A} \textbf{37.09}} & {\color[HTML]{2C3A4A} \textbf{24.46}} \\ \midrule
                          & \multicolumn{3}{c}{\cellcolor[HTML]{FFFFFF}50K Cases}                                                                 \\ \midrule
ModernBert                & {\color[HTML]{333333} 4.94}           & {\color[HTML]{333333} 15.24}          & {\color[HTML]{333333} 9.58}           \\
GuRE (10K) + ModerBert    & {\color[HTML]{2C3A4A} 8.67}           & {\color[HTML]{2C3A4A} 23.94}          & {\color[HTML]{2C3A4A} 15.53}          \\
GuRE (100 K) + ModernBert & {\color[HTML]{2C3A4A} \textbf{10.3}}  & {\color[HTML]{2C3A4A} \textbf{26.66}} & {\color[HTML]{2C3A4A} \textbf{17.71}} \\ \bottomrule
\end{tabular}%
}
\caption{Retrieval results of GuRE trained under data-scarce settings. GuRE with only 10K training examples outperforms retriever fine-tuning approaches that require millions of examples across all retrieval pools.}
\label{tab:new}
\end{table}
\subsection{Generalizability under Data Constraints}
Although GuRE is designed as a plug-and-play, retriever-agnostic approach, it still requires training. To assess its applicability in data-scarce environments, such as legal systems where case law is only partially available, we conducted experiments with varying training sizes. Results show that GuRE trained on only 10K cases already outperforms retriever fine-tuning across all retrieval pool settings. When trained on 100K cases—a scale more realistic for practical deployment—performance further improves. These findings demonstrate that GuRE remains robust under limited-resource conditions and holds strong potential for practical use across diverse legal systems.

\begin{figure}
    \centering
    \includegraphics[width=0.9\columnwidth]{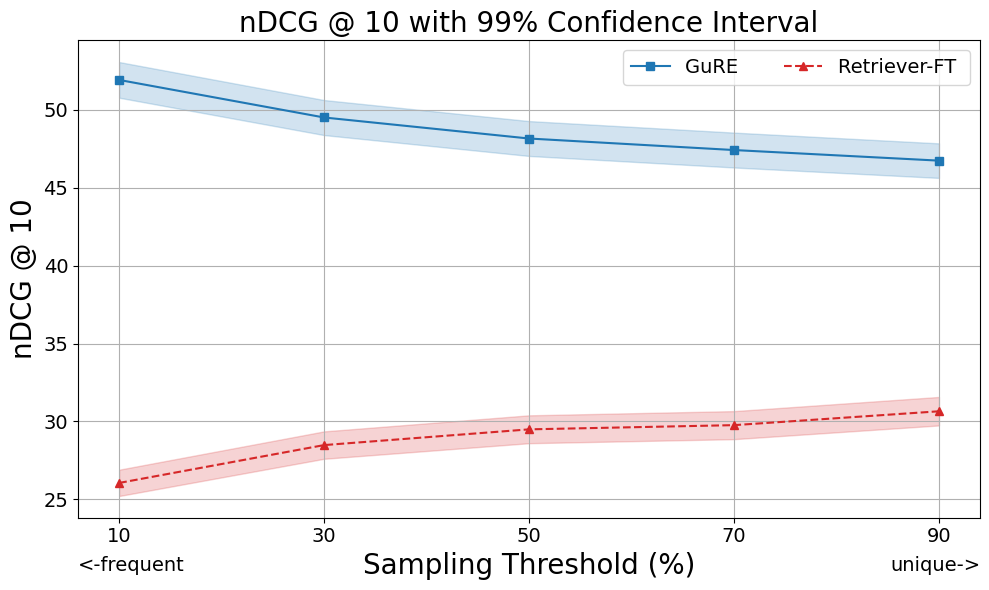}
    \caption{nDCG@10 with 99\% confidence intervals (shading) for GuRE and a fine-tuned retriever across sampling thresholds. 
    Higher thresholds yield more unique samples, while lower ones favor frequent samples.
    Retriever for this experiment is ModernBert.}
    \label{fig:3}
\end{figure}

\subsection{Which Model Should We Train?} \vspace{-7pt} \label{loss}
Citations in U.S. federal precedents follow a long-tailed distribution, with the top 1\% of passages accounting for 18\% of all citations, while 64\% receive only one citation \cite{mahari-etal-2024-lepard}. To investigate the impact of this imbalance, we analyze performance changes by varying the frequency thresholds of test samples. We sort test candidates (10\%) by their frequency in the training set (90\%) and select from the top X\% most frequent passages (X = 10, 30, 50, 70, 90) from test candidates. As X increases, the test set includes more unique passages. We sample 10,000 examples per threshold.

Figure \ref{fig:3} shows that GuRE consistently outperforms fine-tuned retrievers at every threshold. 
Notably, while the performance of GuRE improves as the samples become frequent, the fine-tuned retriever shows the opposite trend. 
This tendency seems to arise from the learning objective used in retriever fine-tuning, which treats all samples in the batch, except the current one, as negative.
In a long-tail distribution, frequent samples appear more frequently in the batch and should be treated as positive since they refer to identical passages. 
However, widely used retriever training losses that rely on in-batch negatives treat them as negative samples.
This may hinder ideal optimization and lead to suboptimal results.
Thus, \textbf{GuRE may be more suitable for LPR, where frequently cited passages are repeatedly referenced.}
More analysis about loss functions is in the Appendix \ref{appendixf}.

\section{Conclusion}
We propose GuRE, a retriever-agnostic query rewriter that mitigates vocabulary mismatch through domain-specific query rewriting.
Experimental results show that GuRE outperforms all baseline methods, including fine-tuned retrievers.
Our analysis highlights why retriever fine-tuning relying on in-batch negatives leads to suboptimal performance in LPR, linking to its loss function.
\section*{Limitations}
\paragraph{Limited Scope} Our experiments are limited to a U.S. federal court precedents-based dataset (LePaRD), which is the only publicly available LPR dataset to our knowledge. 
In the future, we hope to expand this work with more diverse resources, including multilingual and cross-jurisdictional applications.

\paragraph{High Computational Resource} Although GuRE significantly outperforms other baseline methods, GuRE also incurs higher computational costs during training, requiring about twice the GPU hours compared to direct retriever training. 
However, once trained, it can be used as a plug-in for any retriever without further fine-tuning, unlike retrievers that require separate training per model. Details are in Appendix \ref{appendixe}

\section*{Ethical Considerations}
\paragraph{Offensive Language Warning}
The dataset used in this study includes publicly available judicial opinions, which may contain offensive or insensitive language. 
Users should be aware of this when interpreting the results.

\paragraph{Data Privacy}
The dataset used in this study consists of publicly available textual data provided by Harvard’s Case Law Access Project (CAP).
Our work does not involve user-related or private data that is not publicly available.

\paragraph{Intended Use}
This work introduces a methodology for legal passage retrieval and is not intended for direct use by individuals involved in legal disputes without professional assistance.
Our approach aims to advance legal NLP research and could support real-world systems that assist legal professionals. 
We hope such technologies improve access to legal information.

\paragraph{License of Artifacts}
This research utilizes Meta Llama 3, licensed under the Meta 
\href{https://www.llama.com/llama3/license/}{Llama 3 Community License} (Copyright © Meta Platforms, Inc.).
All other models and datasets used in this study are publicly available under permissive licenses.
\section*{Acknowledgments}
This research was supported by the MSIT(Ministry of Science and ICT), Korea, under the ITRC(Information Technology Research Center) support program(IITP-2025-RS-2020-II201789) supervised by the IITP(Institute for Information \& Communications Technology Planning \& Evaluation, Contribution Rate: 45\%). This research was supported by Culture, Sports and Tourism R\&D Program through the Korea Creative Content Agency grant funded by the Ministry of Culture, Sports and Tourism in 2025 (Project Name: Development of an AI-Based Korean Diagnostic System for Efficient Korean Speaking Learning by Foreigners, Project Number: RS-2025-02413038, Contribution Rate: 45\%). This work was also supported by the Institute of Information \& Communications Technology Planning \& Evaluation (IITP) grant funded by the Korea government (MSIT) (No. RS-2019-II191906, Artificial Intelligence Graduate School Program (POSTECH), Contribution Rate: 10\%).

\bibliography{custom}

\appendix
\newpage
\section{Details of Baselines} \label{appendix}
\paragraph{Vanilla Retriever}
Given an ongoing context \( q \), the retriever retrieves the most relevant passage from the candidate set \( P_{collection} \). This approach directly uses \( q \) without any modification.

\paragraph{Query2Doc}  
Query2Doc \cite{wang-etal-2023-query2doc} (Q2D) generates a pseudo-passage via few-shot prompting and concatenates it with the original query to form an expanded query. More formally:
\[
q^+ = \text{concat}(q, \text{LLM}(\text{Prompt}_q))
\]
$\text{LLM(Prompt}_q\text{)}$ represent generated pseudo passage from few-shot Q2D prompt.
Q2D uses $q^+$ to retrieve the most relevant passage.

\paragraph{Query2Doc-CoT}  
Query2Doc-CoT \cite{jagerman2023query} (Q2D-CoT) extends Query2Doc by generating reasoning steps before producing the pseudo-passage. More formally:
\[
q^+ = \text{concat}(q, \text{LLM}(\text{CoTPrompt}_q))
\]
$\text{LLM(CoTPrompt}_q\text{)}$ represent generated pseudo passage from few-shot Q2D-CoT prompt.
Q2D-CoT uses $q^+$ to retrieve the most relevant passage, similar to the approach used by Q2D.

\paragraph{Retrieval Fine Tuning}
We directly train retrieval models using Multiple Negatives Ranking Loss \cite{henderson2017efficient}, where the model is optimized to maximize similarity for positive samples within a batch while minimizing similarity for other negative samples.
The loss is defined as:
\[
\mathcal{L} = -\log \frac{e^{\text{sim}(q, p^+)}}{e^{\text{sim}(q, p^+)} + \sum_{i=1}^{N} e^{\text{sim}(q, p^-_i)}}
\]
\( \text{sim}(q, p) \) represents the similarity score.
Here, $q$ denotes the query, $p^+$ is the positive passage, and $p^-$ refers to other passages in the same batch.

\section{Detailed Dataset Statistics} \label{appendixa}
LePaRD \cite{mahari-etal-2024-lepard} captures citation relationships in U.S. federal court precedents, reflecting how judges use precedential passages based on millions of decisions. 
As shown in Table \ref{tab:5}, the dataset has three versions, each with a different number of target passages in the retrieval pool. 
Each data point pairs a passage before a precedent’s citation with its citation.

\begin{figure}[ht]
    \centering
    \includegraphics[width=\columnwidth]{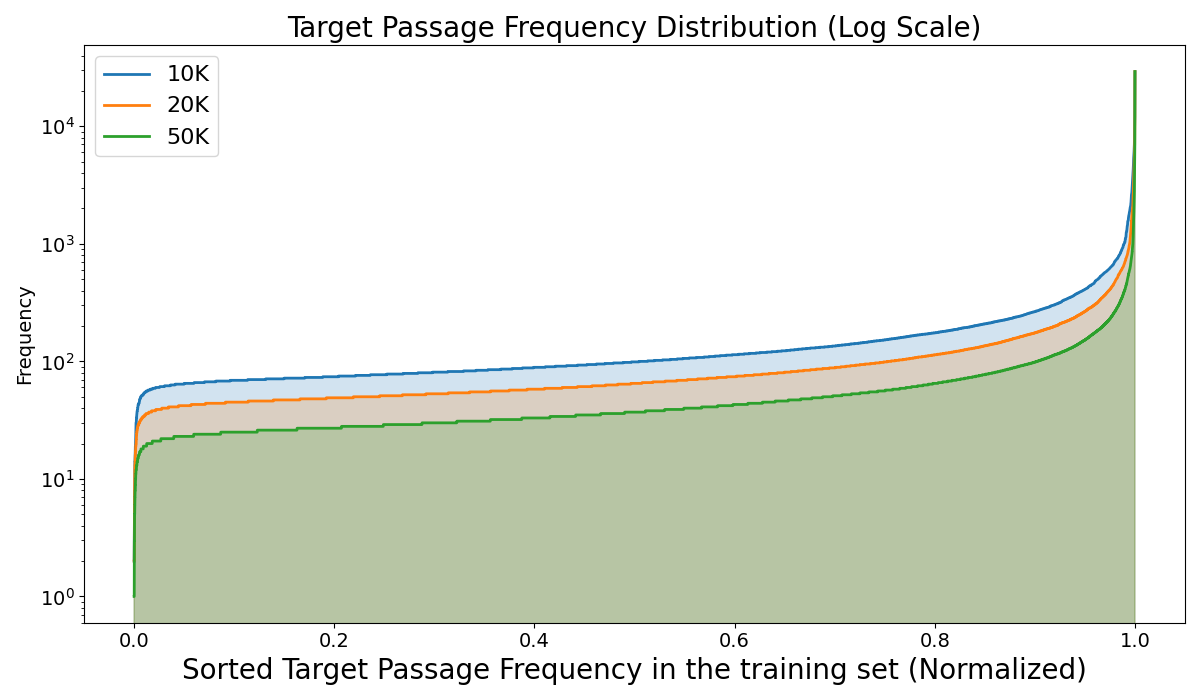}
    \caption{Target passage frequency distribution across different dataset versions (Log Scale)}
    \label{fig:fig4}
\end{figure}

The dataset follows a long-tailed distribution, where the top 1\% of passages (100, 200, or 500) account for 16.23\% to 16.86\% of the data, indicating dominance by a small number of heavily cited precedents.
This tendency is further evident in the dataset distribution visualized in Figure \ref{fig:fig4}. Despite being plotted on a log scale, the distribution shows a remarkable long-tail pattern, where an extremely small number of passages dominate the dataset.
\begin{table}[ht]
\centering
\resizebox{\columnwidth}{!}{%
\begin{tabular}{@{}lccc@{}}
\toprule
\multicolumn{1}{c}{Number of target passages} & Total  & Train (90\%) & Top 1\% population \\ \midrule
10,000 (10 K)                                 & 1.92 M & 1.73 M       & 16.86 \%           \\
20,000 (20 K)                                 & 2.48 M & 2.23 M       & 16.45 \%           \\
50,000 (50 K)                                 & 3.50 M & 3.15 M       & 16.23 \%           \\ \bottomrule
\end{tabular}%
}
\caption{Detailed statistics of LePaRD dataset. }
\label{tab:5}
\end{table}

\section{Query Expansion \& Rewriting Details} \label{appendixb}
\subsection{Prompts} 
\begin{figure}[ht]
\centering
\footnotesize
\textbf{Q2D Prompt}
\begin{verbatimbox}
Write a following legal passage that is coherent, relevant, and contextually appropriate based on preceding context.

Examples:

\#\#\# Preceding Context : \{Example Context 1\}

\#\#\# Legal Passage : \{Example Passage 1\}

\#\#\# Preceding Context : \{Example Context12\}

\#\#\# Legal Passage : \{Example Passage 2\}

\#\#\# Preceding Context : \{Example Context 3\}

\#\#\# Legal Passage : \{Example Passage 3\}

Query:

\#\#\# Preceding Context : \{Context\}

\#\#\# Legal Passage :
\end{verbatimbox}
\caption{Q2D prompt}
\label{fig:4}
\end{figure}

\paragraph{Q2D Prompt} Figure \ref{fig:4} illustrates the prompt used for the Query2Doc\cite{wang-etal-2023-query2doc} method in our experiment. 
As introduced in Query2Doc, we adapt a few-shot prompting paradigm to generate the pseudo-passage, which we adapt to suit legal passage retrieval. 
We randomly select three data points from the training set for the experiment and employ them as fixed examples in the prompt. 
Due to the long length of the actual examples, we replace them with placeholders in Figure \ref{fig:4}.

\begin{figure}[ht]
\centering
\footnotesize
\textbf{Q2D-CoT Prompt}

\begin{verbatimbox}
Write a following legal passage that is coherent, relevant, and contextually appropriate based on preceding context.

\#\#\# Note: Examples provided below do not include intermediate steps due to sampling constraints.

\#\#\# Step 1: Understand the preceding context.

\#\#\# Step 2: Identify the key legal elements and principles required for coherence.

\#\#\# Step 3: Generate a legal passage that logically follows and aligns with the context.

\#\#\# Note: You can generate any intermediate step but, please mark final output with '<output>' tag.

Examples:

\#\#\# Preceding Context : \{Example Context 1\}

\#\#\# Step1: \{Example1:generated step 1\}

\#\#\# Step2: \{Example1:generated step 2\}

\#\#\# Step3: <output> \{Example Passage 1\}

\#\#\# Preceding Context : \{Example Context 2\}

\#\#\# Step1: \{Example2:generated step 1\}

\#\#\# Step2: \{Example2:generated step 2\}

\#\#\# Step3: <output> \{Example Passage 2\}

\#\#\# Preceding Context : \{Example Context 3\}

\#\#\# Step1: \{Example3:generated step 1\}

\#\#\# Step2: \{Example3:generated step 3\}

\#\#\# Step3: <output> \{Example Passage 3\}

Query:

\#\#\# Preceding Context : \{Context\}

\#\#\# Legal Passage :
\end{verbatimbox}
\caption{Q2D-CoT prompt}
\label{fig:5}
\end{figure}

\paragraph{Q2D-CoT Prompt} Figure \ref{fig:5} illustrates the prompt used for the Q2D-CoT\cite{jagerman2023query} method in our experiment. Like Query2Doc, we adapt the few-shot prompting paradigm to suit our task of legal passage retrieval.We randomly select three data points from the training set and use them as fixed examples in the prompt. For the intermediate reasoning steps, we use the zero-shot output from the Q2D-CoT prompt fed into o1 \cite{jaech2024openai}, as shown in Figure \ref{fig:5}.
\begin{table}[ht]
\centering
\resizebox{0.7\columnwidth}{!}{%
\begin{tabular}{@{}llll@{}}
\toprule
         & R@1           & R@10           & nDCG@10        \\ \midrule
Q2D      & \textbf{$\text{6.92}^{\dagger}$} & 24.96          & 14.80          \\
Q2D-TOP3 & 6.15          & \textbf{$\text{27.41}^{\dagger}$} & \textbf{15.44} \\ \bottomrule
\end{tabular}%
}
\caption{Evaluation results on 10,000 samples from 10K dataset by varying in-context example selection methods.
$\dagger$ indicates a statistically significant values (paired $t$-test $p < 0.01$)}
\label{tab:13}
\end{table}
\paragraph{In-context Example Selection}
For the experiment, we randomly select three data points from the training set as fixed examples in the prompt following \citet{wang-etal-2023-query2doc}.
However, some studies suggest that providing pseudo-relevant examples as in-context examples can improve performance \cite{AZAD2022148,jagerman2023query}.
To investigate this, we conduct a comparative analysis of in-context example selection methods.
We give Top-3 relevant examples retrieved by BM25 using query from training set for Q2D-TOP3.

Table \ref{tab:13} compares in-context example selection methods. While Q2D-TOP3 uses pseudo-relevant examples, its advantage is limited to R$@$10, suggesting that \textbf{example selection methods do not significantly impact performance}. So, we use fixed random examples following \cite{wang-etal-2023-query2doc}.

\subsection{Decoding} \label{appendixb2}
We apply nucleus decoding \cite{DBLP:conf/iclr/HoltzmanBDFC20} for the baselines and GURE, with a temperature of 0 and a top-p value of 0.9. 
GuRE takes approximately 10 to 12 minutes to generate 10,000 samples using vLLM \cite{10.1145/3600006.3613165} on an NVIDIA RTX 3090 GPU. 
\textbf{This demonstrates that our approach can improve performance with minimal additional latency, under 0.1 seconds per query.}

For the Q2D and Q2D-CoT experiments, we utilize an OpenAI API.
We employ GPT-4o-mini \cite{openai2024gpt4technicalreport}.
The same decoding parameters with GuRE are applied across both methods. 
The total cost for these experiments is \$52.83.

\section{Impact of Backbone Model} \label{appendixbackbone}

\begin{table}[ht]
\centering
\resizebox{0.85\columnwidth}{!}{%
\begin{tabular}{@{}cccc@{}}
\toprule
                    & R @ 1 & R @ 10 & nDCG @ 10 \\ \midrule
\small GuRE (SaulLM-7B)    & \textbf{33.14}  & \textbf{60.24}   & \textbf{45.86}    \\
\small GuRE (Qwen2.5-7B)   & 26.14  & 51.88   & 38.08    \\
\small GuRE (llama3.1-8B)  & 22.93  & 47.99   & 34.47    \\ \midrule
\small LegalBert-FT  & \textbf{15.35}  & \textbf{56.77}   & \textbf{33.52}    \\
\small ModernBert-FT  & 14.12  & 51.34   & 30.50    \\  \bottomrule
\end{tabular}%
}
\caption{Comparison of LPR results on the 10k test set by varying backbone model of GuRE. We employ vanilla ModernBERT as a retriever for GuRE.}
\label{tab:2}
\end{table}
Table \ref{tab:2} shows that GuRE performs better with legally pre-trained LLMs than with generally pre-trained ones.
GuRE (SaulLM-7B) achieves an R@1 score of 33.14 and nDCG@10 of 45.86,  while GuRE with generally pre-trained LLMs shows suboptimal performance. 
Although GuRE tends to outperform retriever fine-tuning, a similar trend is observed in retriever fine-tuning, where the legally pre-trained LegalBert outperforms one of the most robust retriever models, ModernBert.
This indicates that \textbf{the performance of training-based methods is impacted by the underlying domain-specific knowledge of the backbone model.}

\section{Details on Retrievers} \label{appendixc}
Dense retrievers encode queries into embedding vectors and retrieve passages based on their cosine similarity in the embedding space.

\paragraph{BM25} BM25 \cite{robertson2009probabilistic} is a sparse retriever based on term frequency-inverse document frequency (TF-IDF). We use BM25s \cite{bm25s} Python library for indexing and retrieval.  

\paragraph{DPR} DPR \cite{karpukhin-etal-2020-dense} is a dense retrieval model that encodes queries and passages into dense vectors. We use DPR\footnote{\href{https://huggingface.co/sentence-transformers/facebook-dpr-ctx_encoder-multiset-base}{sentence-transformers/facebook-dpr-ctx\_encoder-multiset-base}} with Sentence Transformers \cite{reimers-2019-sentence-bert}.  

\paragraph{ModernBERT} ModernBERT\footnote{\href{https://huggingface.co/Alibaba-NLP/gte-modernbert-base}{Alibaba-NLP/gte-modernbert-base}} \cite{warner2024smarterbetterfasterlonger} achieves state-of-the-art performance in single- and multi-vector retrieval across domains. 
We use it similarly to DPR, encoding text into embeddings for retrieval.

\paragraph{LegalBERT}  
LegalBERT\footnote{\href{https://huggingface.co/nlpaueb/legal-bert-base-uncased}{nlpaueb/legal-bert-base-uncased}} \cite{chalkidis-etal-2020-legal} is trained from scratch on a large corpus of legal documents. 
Since LegalBert is not pre-trained to produce sentence embedding vectors, we do not use it directly for dense retrieval, instead fine-tune it for downstream tasks.

\section{Evaluation Metrics}

\paragraph{Retrieval}
We evaluate the performance of our retrievers using 
Recall$@$1, Recall$@$10, nDCG$@$10.
\textbf{Recall$@$1} measures the proportion of queries for which the correct passage is ranked first in the retrieved list.
\textbf{Recall$@$10} extends this by measuring the proportion of queries for which the correct passage appears in the top 10 retrieved passages. It reflects the model’s ability to identify relevant passages within a broader set of candidates.
\textbf{nDCG$@$10}  (Normalized Discounted Cumulative Gain at 10) considers the position of relevant passages, giving higher weight to passages ranked closer to the top.

\paragraph{Generation}
For quantitative evaluation of generated pseudo passages, we use BLEU \cite{papineni-etal-2002-bleu}, ROUGE-L \cite{lin-2004-rouge} and BertScore-F \cite{bert-score}.
\textbf{BLEU} measures the precision of n-grams between the generated text and the reference text. It evaluates how much of the generated text matches the reference, with a higher score indicating better accuracy of the generated text.
\textbf{ROUGE-L} focuses on the longest common subsequence between the generated and reference texts. It emphasizes the recall aspect of the overlap.
\textbf{BertScore-F} evaluates the similarity between generated and reference texts using contextual embeddings from BERT. A higher score indicates that the generation closely aligns with the reference's meaning.

\section{Training Details} \label{appendixe}
\paragraph{Retriever}
For training the dense retrievers, we utilized implemented libraries: the Sentence Transformers\cite{reimers-2019-sentence-bert} and accelerate \cite{accelerate}. 
The training was conducted with a batch size of 32 per device, over 3 epochs, with a maximum sequence length of 256. 
The warm-up step ratio was set to 0.1. We utilized the Multiple Negative Ranking Loss function for training as mentioned in the main text.
We trained the model using RTX 3090 GPUs. The training time varied depending on the dataset size:20 GPU hours for 10K, 30 GPU hours for 20K, 44 GPU hours for 50K dataset.

\paragraph{GuRE}
For training GuRE, we utilized transformers \cite{wolf-etal-2020-transformers}, Trl \cite{vonwerra2022Trl}, deepspeed \cite{10.1145/3394486.3406703}, and accelerate. 
The model was trained with a LoRA \cite{hu2022lora} rank of 64, a cosine learning rate scheduler, and the AdamW \cite{loshchilov2018decoupled} optimizer over 1 epoch. 
The per-device batch size was set to 4, and the learning rate was 5e-5. 
We used the SFT trainer from Trl for training.
We trained the model using RTX A6000 GPUs and RTX 6000ADA GPUs. The training time varied depending on the dataset size: 60 GPU hours for the 10K, 100 GPU hours for the 20K, and 130 GPU hours for the 50K dataset.

While training the GuRE model takes more GPU hours than direct retriever fine-tuning, it offers significant advantages. 
GuRE can be applied in a \textbf{retriever-agnostic manner once trained, making it a more efficient solution.}

\section{Analysis on Training Objectives} \label{appendixf}
We chose Multiple Negative Ranking Loss (MNRL) due to the large dataset scale, where explicit negative sampling is costly. Since each query only matches one positive passage, MNRL was effective in this setup.

However, as seen in Table \ref{tab:5} and Figure \ref{fig:fig4} , the dataset is dominated by a small number of heavily cited precedents. 
Frequent samples, though positive, are treated as negative by the model, leading to reduced accuracy in these passages. 
This is problematic because frequently cited precedents are crucial in legal cases, and lower accuracy on them reduces the system's practical usefulness.

\subsection{Trade-Off in Reducing In-Batch Negative Sensitivity} To reduce this in-batch negative sensitivity, we experimented with a contrastive loss that is unaffected by in-batch samples.
\[
L = \frac{1}{2} \left( y \cdot D^2 + (1 - y) \cdot \max(0, m - D)^2 \right)
\]
Here, $y$ represents the label, where 1 for positive passages and 0 for negative passages. $D$ is the distance between the query and the passage in the embedding space, and $m$ is the margin.
For positive pairs, the loss encourages the distance $D$ to be small, while for negative pairs, the loss pushes the distance $D$ to be larger than the margin $m$.

For each query, we formed positive and negative triples by pairing the query with its corresponding target passage and a hard negative, which was the highest-ranked passage from the BM25 results that was not the target passage.
\begin{table}[ht]
\centering
\resizebox{\columnwidth}{!}{%
\begin{tabular}{@{}ccccc@{}}
\toprule
     & R@1   & R@10  & nDCG@10 & GPU hours \\ \midrule
MNRL & 13.28 & 48.86 & 28.92   & 20        \\
CL   & 0.1   & 0.65  & 0.34    & 30        \\ \bottomrule
\end{tabular}%
}
\caption{Retrieval performance on the 10K dataset using ModerBERT trained with Multiple Negative Ranking Loss (MNRL) and Contrastive Loss (CL). CL requires explicit negative samples, increasing GPU training time as the number of negatives grows. In contrast, MNRL relies on in-batch negative samples, making GPU hours dependent on batch size.}
\label{tab:6}
\end{table}

However, the model's performance dropped significantly compared to MNRL, as shown in Table \ref{tab:6}. While MNRL learns from (batchsize - 1) negative samples, contrastive loss only considers a limited number of explicitly labeled hard negative samples.
\textbf{Nevertheless, increasing the number of negative samples for exposing various negative samples like MNRL would require significantly more training time, making it inefficient and impractical for large-scale applications.}
Therefore, as discussed in Section \ref{loss}, GuRE proves to be more effective for real-world scenarios, offering a more efficient approach.

\subsection{Supplementary Graphs} 
\begin{figure}[ht]
    \centering
    \includegraphics[width=0.5\columnwidth]{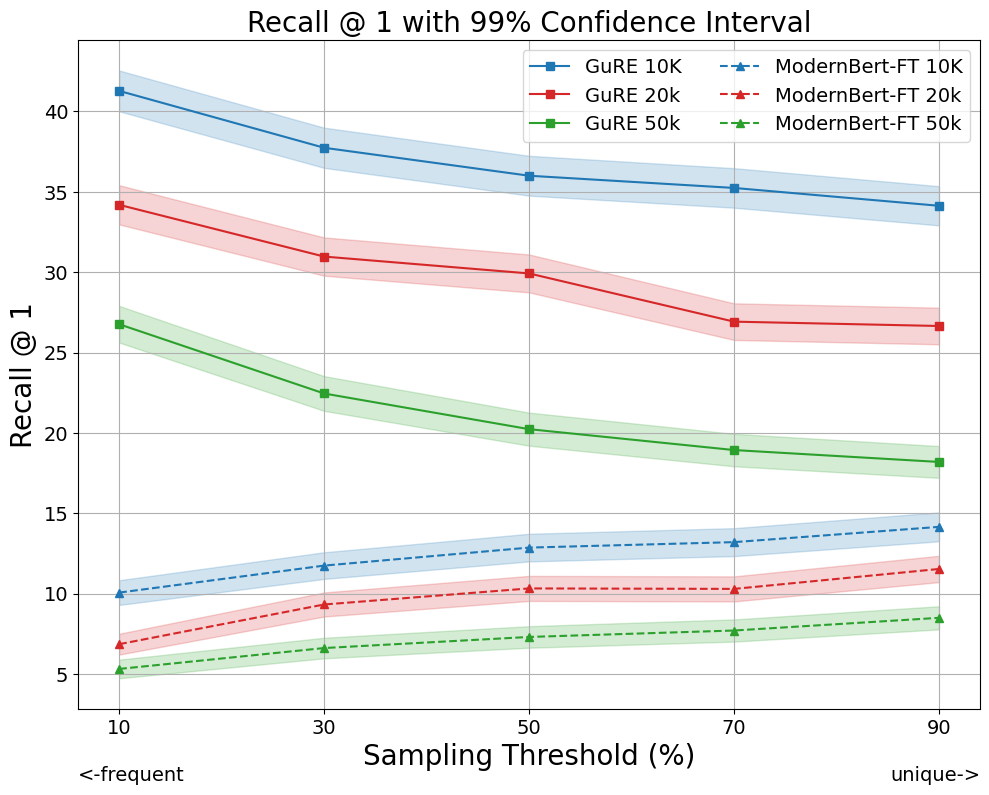}
    \caption{Recall@1 with 99\% confidence intervals (shading) for GuRE and a fine-tuned retriever across sampling thresholds. 
}
    \label{fig:r1}
\end{figure}

\begin{figure}[ht]
    \centering
    \includegraphics[width=0.5\columnwidth]{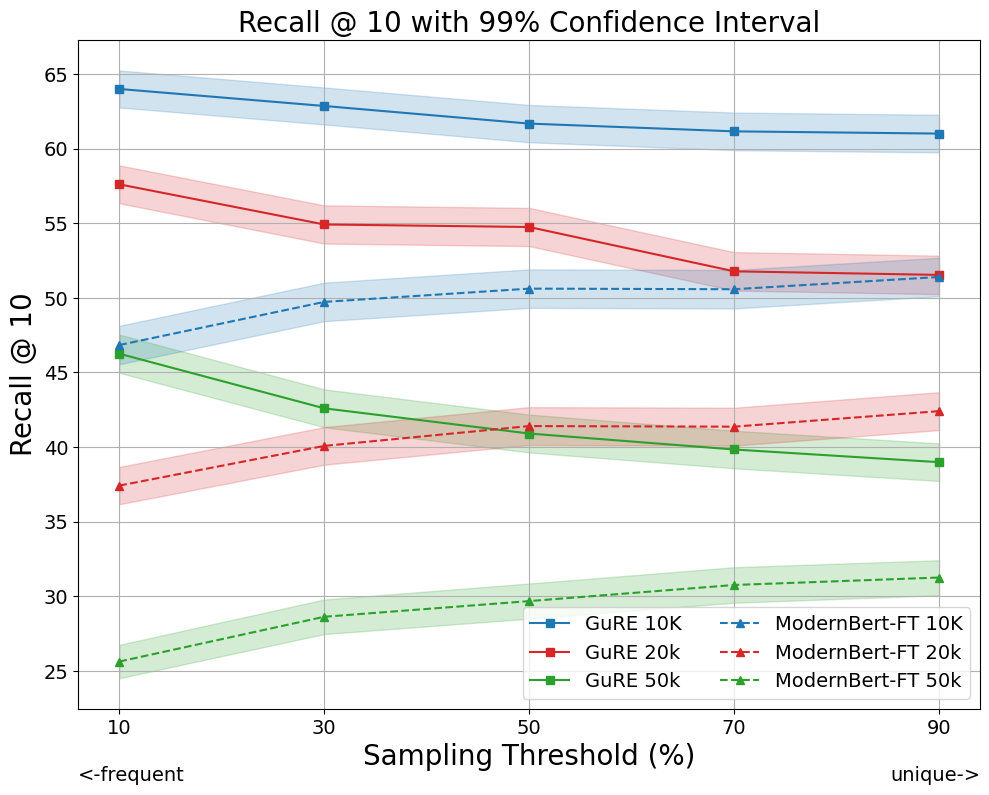}
    \caption{Recall@10 with 99\% confidence intervals (shading) for GuRE and a fine-tuned retriever across sampling thresholds.}
    \label{fig:r10}
\end{figure}

\begin{figure}[ht]
    \centering
    \includegraphics[width=0.5\columnwidth]{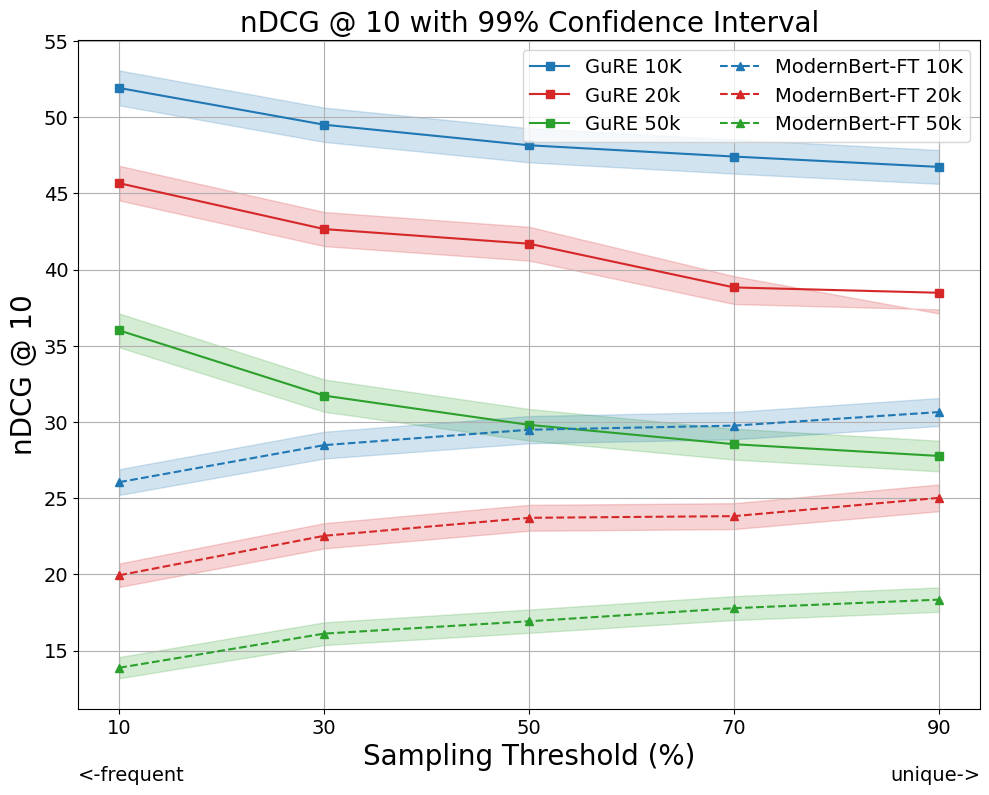}
    \caption{nDCG@10 with 99\% confidence intervals (shading) for GuRE and a fine-tuned retriever across sampling thresholds.}
    \label{fig:ndcg}
\end{figure}

The Figures (\ref{fig:r1}, \ref{fig:r10} ,\ref{fig:ndcg}) show performance across different frequency thresholds for various data versions, supplementing Figure \ref{fig:3} in the main body. 
As seen in the figures, the performance trend based on the training objective is consistent across all datasets and metrics.
Higher thresholds yield more unique samples, while lower ones favor frequent samples.
Retriever for this experiment is ModernBert.

\section{Case Studies} \label{appendixg}
We conduct a case study to better understand the impact of the baseline methods and GuRE on the retriever. 
The following tables show the query and the top 5 retrieval results, varying by method.

Other baseline methods struggle to retrieve the target passage due to vocabulary mismatches between the query and the target passage (Table \ref{tab:7}, \ref{tab:10}) or because the expanded query includes irrelevant information which may incur hallucination problems mentioned in Introduction (Table  \ref{tab:8}, \ref{tab:9}).
 However, GuRE generates a query identical to the target passage (Table \ref{tab:11}).

\onecolumn

\newpage

\begin{table*}[t]
\centering
\resizebox{\textwidth}{!}{%
\begin{tabular}{@{}p{0.15\textwidth}p{0.85\textwidth}@{}}
\toprule[1.5pt]
\small Query & \small Horowitz v. Fed. Kemper Life Assurance Co., 57 F.3d 300, 302 n. If the moving party has demonstrated \highlightpink{an absence of material fact,} the nonmoving party then \highlightpink{“must come forward with ‘specific facts showing that there is a genuine issue for trial.}  \\ \hline \midrule
\small Target Passage   &  \small \highlightcyan{In the language of the Rule, the nonmoving party must come forward with “specific facts showing that there is a genuine issue for trial.}  \\ \hline \midrule
\small Rank 1   &  \small So long as the movant has met its initial burden of “demonstrat[ing] the absence of a genuine issue of material fact,” id. at 323, 106 S.Ct. at 2553, the nonmoving party then \highlightpink{“must set forth specific facts showing that there is a genuine issue for trial.} \\  \midrule
\small Rank 2   &  \small We will affirm if the evidence, viewed in the light most favorable to the nonmoving party, shows that \highlightpink{no genuine issue of material fact} exists and that the moving party is entitled to judgment as a matter of law. \\  \midrule
\small Rank 3  &  \small Once the non-movant demonstrates that there is an absence of evidence to support the non-moving party’s case, the movant must then \highlightpink{set forth specific facts showing that there is an issue for trial.}  \\ \midrule
\small Rank 4    &  \small In such a situation, there can be \highlightpink{“no genuine issue as to any material fact,”} since a complete failure of proof concerning an essential element of the nonmoving party’s case necessarily renders all other facts immaterial. \\ \midrule
\small Rank 5    &  \small  Once the moving party has initially shown “that there is an absence of evidence to support the non-moving party’s cause,” Celotex Corp. v. Catrett, 477 U.S. 317, 325, 106 S.Ct. 2548, 91 L.Ed.2d 265 (1986), the non-movant \highlightpink{must come forward with “specific facts” showing a genuine factual issue for trial.} \\  \bottomrule[1.5pt]
\end{tabular}%
}
\caption{Top-5 Retrieval results using vanilla ModerBert and query without any modification. \highlightcyan{Cyan} indicates the target passage and the correct answer among candidates.
\highlightpink{Pink} indicates potential \textit{"distractor"} that can mislead retrievers into selecting an irrelevant passage.
In this case, the retriever fails to include the correct passage due to the vocabulary mismatch between the query and the target passage.}
\label{tab:7}
\end{table*}
\begin{table*}[t]
\centering
\resizebox{\textwidth}{!}{%
\begin{tabular}{@{}p{0.15\textwidth}p{0.86\textwidth}@{}}
\toprule[1.5pt]
\small Query (Query + pseudo-passage)& \small (Original query) Horowitz v. Fed. Kemper Life Assurance Co., 57 F.3d 300, 302 n. If ... (skip) + \highlightyellow{[W]hen evaluating whether a genuine issue exists, the court must consider the evidence in the light most favorable to the nonmoving party, drawing all reasonable inferences in their favor, while disregarding any evidence that constitutes mere speculation or conjecture. This standard underscores the principle that summary judgment is not to be used as a substitute for a trial, but rather as a mechanism to resolve disputes where} \highlightpink{no material facts are genuinely in contention.} \\ \hline \midrule
\small Target Passage   &  \small \highlightcyan{In the language of the Rule, the nonmoving party must come forward with “specific facts showing that there is a genuine issue for trial.}  \\ \hline \midrule
\small Rank 1   & \small The court noted that on a motion for summary judgment, once the moving party has proffered facts to show that there is \highlightpink{no genuine issue as to any material fact} and that that party is entitled to judgment as a matter of law, “the opposing party must present ‘specific facts showing [that] there is a genuine issue for trial.  \\ \midrule
\small Rank 2     & \small  We will affirm if the evidence, viewed in the light most favorable to the nonmoving party, shows that \highlightpink{no genuine issue of material fact} exists and that the moving party is entitled to judgment as a matter of law.  \\ \midrule
\small Rank 3  &   \small [T]he mere existence of some alleged factual dispute between the parties will not defeat an otherwise properly supported motion for summary judgment; the requirement is that there be no genuine issue of material fact. \\  \midrule
\small Rank 4    & \small Summary judgment is appropriate where \highlightpink{“there is no genuine issue as to any material fact} and... the moving party is entitled to a judgment as a matter of law,” Fed.R.Civ.P. 56(c), i.e., “[w]here the record taken as a whole could not lead a rational trier of fact to find for the non-moving party. \\ \midrule
\small Rank 5    & \small A motion for summary judgment should be granted if, viewing the evidence in the light most favorable to the nonmoving party, \highlightpink{‘there is no genuine issue as to any material fact} and if the moving party is entitled to judgment as a matter of law. \\ \bottomrule[1.5pt]
\end{tabular}%
}
\caption{Top-5 Retrieval results using vanilla ModerBert and a pseudo-passage generated through Q2D.
\highlightyellow{Yellow} indicates generated context from Q2D.
\highlightcyan{Cyan} indicates target passage.
\highlightpink{Pink} indicates potential \textit{"distractor"} that can mislead retrievers into selecting an irrelevant passage.
In this case, the retriever fails to include the correct passage due to the generated irrelevant context.}
\label{tab:8}
\end{table*}
\begin{table*}[t]
\centering
\resizebox{\textwidth}{!}{%
\begin{tabular}{@{}p{0.15\textwidth}p{0.86\textwidth}@{}}
\toprule[1.5pt]
\small Query (Query + pseudo-passage)& \small (Original query) Horowitz v. Fed. Kemper Life Assurance Co., 57 F.3d 300, 302 n. If ... (skip) + \highlightyellow{the nonmoving party must set forth specific facts demonstrating that genuine issues exist for trial.}  \\ \hline \midrule
\small Target Passage   &   \small \highlightcyan{In the language of the Rule, the nonmoving party must come forward with “specific facts showing that there is a genuine issue for trial.}  \\ \hline \midrule
\small Rank 1   &  \small So long as the movant has met its initial burden of “demonstrat[ing] the absence of a genuine issue of material fact,” id. at 323, 106 S.Ct. at 2553, the nonmoving party then \highlightpink{“must set forth specific facts showing that there is a genuine issue for trial.} \\ \midrule
\small \textbf{Rank 2 (Correct)}    &  \small \highlightcyan{In the language of the Rule, the nonmoving party must come forward with “specific facts showing that there is a genuine issue for trial.}  \\ \midrule
\small Rank 3  & \small Although the moving party bears the initial burden of establishing that there are no genuine issues of material fact, once such a showing is made, the non-movant must \highlightpink{“set forth specific facts showing that there is a genuine issue for trial.}   \\ \midrule
\small Rank 4    & \small  The nonmoving party may not, however, “rest on mere allegations or denials” but must demonstrate on the record the existence of \highlightpink{specific facts which create a genuine issue for trial.} \\ \midrule
\small Rank 5    & \small The court noted that on a motion for summary judgment, once the moving party has proffered facts to show that there is no genuine issue as to any material fact and that that party is entitled to judgment as a matter of law, “the opposing party must present \highlightpink{‘specific facts showing [that] there is a genuine issue for trial.} \\ \bottomrule[1.5pt]
\end{tabular}%
}
\caption{Top-5 Retrieval results using vanilla ModerBert and a pseudo-passage generated through Q2D-CoT.
\highlightyellow{Yellow} indicates generated context from Q2D-CoT.
\highlightcyan{Cyan} indicates the target passage and the correct answer among candidates.
\highlightpink{Pink} indicates potential \textit{"distractor"} that can mislead retrievers into selecting an irrelevant passage. In this case, the entire generated query plays the role of a \textit{"distractor"}.}
\label{tab:9}
\end{table*}
\begin{table*}[t]
\centering
\resizebox{\textwidth}{!}{%
\begin{tabular}{@{}p{0.15\textwidth}p{0.86\textwidth}@{}}
\toprule[1.5pt]
\small Query & \small Horowitz v. Fed. Kemper Life Assurance Co., 57 F.3d 300, 302 n. If the moving party has demonstrated \highlightpink{an absence of material fact}, the nonmoving party then “must \highlightpink{come forward with ‘specific facts showing that there is a genuine issue for trial.}  \\ \hline \midrule
\small Target Passage   &   \small \highlightcyan{In the language of the Rule, the nonmoving party must come forward with “specific facts showing that there is a genuine issue for trial.}  \\ \hline \midrule
\small Rank 1   & \small Once the movant, here, the defendant, satisfies its initial burden under Rule 56(c) of demonstrating \highlightpink{the absence of a genuine issue of material fact}, the burden shifts to the nonmovant to \highlightpink{“come forward with ‘specific facts showing that there is a genuine issue for trial.}  \\ \midrule
\small Rank 2     &  \small To do so successfully, the non-moving party must demonstrate more than “some metaphysical doubt as to the material facts,... [it] must \highlightpink{come forward with ‘specific facts showing that there is a genuine issue for trial.} \\ \midrule
\small Rank 3  & \small The nonmovant “must do more than simply show that there is some metaphysical doubt as to the material facts,” and \highlightpink{must come forward with “specific facts showing that there is a genuine issue for trial.}   \\ \midrule
\small \textbf{Rank 4 (Correct)}   & \small \highlightcyan{In the language of the Rule, the nonmoving party must come forward with “specific facts showing that there is a genuine issue for trial.} \\ \midrule
\small Rank 5    & \small If the movant demonstrates \highlightpink{an absence of a genuine issue of material fact}, a limited burden of production shifts to the non-movant, who must “demonstrate more than some metaphysical doubt as to the material facts,” and come forward with \highlightpink{“specific facts showing that there is a genuine issue for trial.} \\ \bottomrule[1.5pt]
\end{tabular}%
}
\caption{Top-5 Retrieval results using fine-tuned ModerBert and query without any modification.
\highlightcyan{Cyan} indicates target passage.
\highlightpink{Pink} indicates potential \textit{"distractor"} that can mislead retrievers into selecting an irrelevant passage.}
\label{tab:10}
\end{table*}
\begin{table*}[t]
\centering
\resizebox{\textwidth}{!}{%
\begin{tabular}{@{}p{0.15\textwidth}p{0.86\textwidth}@{}}
\toprule[1.5pt]
\small Query (\textit{"rewritten query"}) & \small \highlightyellow{In the language of the Rule, the nonmoving party must come forward with “specific facts showing that there is a genuine issue for trial.}  \\ \hline \midrule
\small Target Passage   &  \small \highlightcyan{In the language of the Rule, the nonmoving party must come forward with “specific facts showing that there is a genuine issue for trial.}  \\ \hline \midrule
\small \textbf{Rank 1 (Correct)}  &  \small \highlightcyan{In the language of the Rule, the nonmoving party must come forward with “specific facts showing that there is a genuine issue for trial.} \\ \midrule
\small Rank 2   &  \small The nonmoving party may not, however, “rest on mere allegations or denials” but must demonstrate on the record the existence of specific facts which create a genuine issue for trial. \\ \midrule
\small Rank 3     & \small Instead, the nonmoving party must set forth, by affidavit or as otherwise provided in Rule 56, “specific facts showing that there is a genuine issue for trial. \\ \midrule
\small Rank 4  & \small To do so successfully, the non-moving party must demonstrate more than “some metaphysical doubt as to the material facts,... [it] must come forward with ‘specific facts showing that there is a genuine issue for trial.  \\ \midrule
\small Rank 5    & \small If the moving party meets this burden, the non-moving party then has the burden to come forward with specific facts showing that there is a genuine issue for trial as to elements essential to the non-moving party’s case. \\ \bottomrule[1.5pt]
\end{tabular}%
}
\caption{Top-5 Retrieval results using vanila ModerBert and \textit{"rewritten query"} generated from GuRE. \highlightyellow{Yellow} indicates generated context from GuRE. GuRE generated the same context as the target passage.
\highlightcyan{Cyan} indicates target passage and the correct answer among candidates.
In this case, generated query from GuRE is identical with target passage.}
\label{tab:11}
\end{table*}

\end{document}